\documentclass[lettersize,journal]{IEEEtran}
\usepackage{amsmath,amsfonts}
\usepackage{algorithmic}
\usepackage{algorithm}
\usepackage{array}
\usepackage[caption=false,font=normalsize,labelfont=sf,textfont=sf]{subfig}
\usepackage{textcomp}
\usepackage{stfloats}
\usepackage{url}
\usepackage{verbatim}
\usepackage{graphicx}
\usepackage{cite}
\usepackage{caption}
\usepackage{colortbl}
\usepackage[table]{xcolor}
\usepackage{booktabs}
\usepackage{hyperref}
\hypersetup{
	colorlinks=true,          
	linkcolor=blue,      
	citecolor=blue,            
}

\begin{document}

\title{Mamba-Driven Topology Fusion for Monocular 3D Human Pose Estimation}

\author{Zenghao Zheng, Lianping Yang, Jinshan Pan, Hegui Zhu
	
\thanks{Corresponding author: Lianping Yang. 

Zenghao Zheng, Lianping Yang, Hegui Zhu are with the College of Sciences, Northeastern University, Shengyang 110819, China (email: 2300178@stu.neu.edu.cn, yanglp@mail.neu.edu.cn, zhuhegui@mail.neu.edu.cn)
	
Lianping Yang, Hegui Zhu are also with Key Laboratory of Differential Equations and Their Applications, Northeastern University, Liaoning Provincial Department of Education

Jinshan Pan is with Nanjing University of Science and Technology, Nanjing, China (email: jspan@njust.edu.cn)}

\thanks{Manuscript received April 19, 2021; revised August 16, 2021.}
}

\markboth{Journal of \LaTeX\ Class Files,~Vol.~14, No.~8, August~2021}%
{Shell \MakeLowercase{\textit{et al.}}: A Sample Article Using IEEEtran.cls for IEEE Journals}

\maketitle

\begin{abstract}
Transformer-based methods for 3D human pose estimation face significant computational challenges due to the quadratic growth of self-attention mechanism complexity with sequence length. Recently, the Mamba model has substantially reduced computational overhead and demonstrated outstanding performance in modeling long sequences by leveraging state space model (SSM). However, the ability of SSM to process sequential data is not suitable for 3D joint sequences with topological structures, and the causal convolution structure in Mamba also lacks insight into local joint relationships. 
To address these issues, we propose the Mamba-Driven Topology Fusion framework in this paper. Specifically, the proposed Bone Aware Module infers the direction and length of bone vectors in the spherical coordinate system, providing effective topological guidance for the Mamba model in processing joint sequences. Furthermore, we enhance the convolutional structure within the Mamba model by integrating forward and backward graph convolutional network, enabling it to better capture local joint dependencies. Finally, we design a Spatiotemporal Refinement Module to model both temporal and spatial relationships within the sequence. Through the incorporation of skeletal topology, our approach effectively alleviates Mamba’s limitations in capturing human structural relationships.
We conduct extensive experiments on the Human3.6M and MPI-INF-3DHP datasets for testing and comparison, and the results show that the proposed method greatly reduces computational cost while achieving higher accuracy. Ablation studies further demonstrate the effectiveness of each proposed module. The code and models will be released.

\end{abstract}


\section{Introduction}

\IEEEPARstart{M}{onocular} 3D human pose estimation has garnered significant attention in the field of computer vision in recent years. The primary objective of this task is to estimate the 3D coordinates of body joints from a given 2D image or video captured from a single perspective. Research in this domain plays a pivotal role in enabling various downstream applications, such as action recognition \cite{luvizon2020multi}, human-computer interaction \cite{munea2020progress}, human body reconstruction \cite{jackson20183d, kanazawa2018end}, and autonomous driving \cite{wiederer2020traffic}.

Methods for single-view 3D human pose estimation can be broadly classified into two categories: end-to-end methods and two-stage methods. End-to-end methods \cite{wang20193d, ma2021context} aim to directly recover 3D joint coordinates from RGB images. In contrast, two-stage methods \cite{zheng20213d, shan2022p, li2022mhformer, zhang2022mixste} first estimate 2D joint coordinates from RGB images and then use a lifting network to map the 2D pose to its 3D counterpart. Given the outstanding performance of 2D pose detectors \cite{2016Stacked, sun2019deep, chen2018cascaded}, two-stage methods often achieve higher accuracy in 3D pose estimation, making them the primary focus of this discussion. However, transforming 2D poses into 3D coordinates presents significant challenges, such as the loss of depth information and issues related to self-occlusion. To address these challenges, convolutional neural network (CNN) \cite{pavllo20193d, zheng2020joint, chen2021anatomy} have been widely used to extract depth features from images. More recently, due to the exceptional success of transformer \cite{vaswani2017attention} in computer vision, transformer-based architectures have been extensively adopted in 3D human pose estimation and demonstrated superior performance compared to CNN-based models. For example, PoseFormer \cite{zheng20213d} was the first to introduce transformer into 3D human pose estimation, exploring both spatial and temporal information in pose sequences. MotionBERT \cite{zhu2023motionbert} designed a dual-stream transformer architecture and achieved SOTA performance by incorporating pre-training techniques. HCT \cite{yin2023multibranch} utilized hierarchical cross transformer to encode high-level temporal, spatial and cross-view information to improve unsupervised 3D HPE. Although these methods achieved promising results, they introduced significant computational and memory overhead due to the quadratic complexity of the self-attention mechanism with respect to the length of pose sequences. Approaches such as dividing time frames \cite{zhong2023frame, wu2021limb} and removing redundant sequences \cite{li2022exploiting, li2024hourglass, zhao2023poseformerv2} can alleviate this issue but fail to fundamentally solve the problem.

The structured state space model (SSM) \cite{gu2021efficiently} has recently demonstrated linear scaling with respect to sequence length and achieved state-of-the-art (SOTA) performance across several modalities, including language, speech, and genomics. Mamba \cite{gu2023mamba} is an improvement over SSM, incorporating an input-dependent selection mechanism and an efficient hardware-aware design. Given the Mamba model's outstanding capabilities in long-sequence modeling, its application to 3D pose estimation tasks, which heavily rely on long pose temporal sequences, is highly effective. Several existing methods that incorporate the Mamba into 3D HPE, such as PoseMamba \cite{huang2024posemamba}, PoseMagic \cite{zhang2024pose}, and HGMamba \cite{cui2025hgmamba}, have achieved competitive performance. However, directly applying the Mamba to capture the spatial structure of human posture is not appropriate. In contrast to the self-attention mechanism, which simultaneously attends to all tokens globally, SSM processes sequential data relying only on tokens preceding the current position. However, the spatial configuration of human joints is inherently non-sequential, as it adheres to the intrinsic skeletal topology of the human body. Moreover, prior to applying SSM, Mamba employs causal convolutions to capture local dependencies within the sequence. This causal convolution strategy is also suboptimal for modeling the complex structural relationships inherent in the human body.

To address the aforementioned limitations, we propose the Mamba-Driven Topology Fusion framework, which integrates skeletal topology to provide structural guidance for Mamba when processing human joint sequences. The framework primarily consists of two components: a bone aware module and a spatiotemporal refinement module. The primary function of the bone aware module is to provide bone direction and length information, thereby offering topological guidance for the spatiotemporal refinement module. Within this module, bone directions are decomposed into azimuth and polar angles, with the polar angle discretized into several categories based on its angular magnitude. To predict these categories, we develop a pyramid-structured classification network built upon the Mamba architecture. Subsequently, the spherical coordinate representation of bone vectors is computed based on the human skeletal topology. 

In the spatiotemporal refinement module, drawing upon the encoder block architecture of Vision Mamba \cite{zhu2024vision}, we propose a upgraded module termed GCN-Enhanced Vision Mamba (GEM). Graph Convolutional Network (GCN) \cite{defferrard2016convolutional} have proven effective for modeling sequential data with inherent graph structures, and have been integrated with Mamba in several recent approaches, including PoseMagic \cite{zhang2024pose}, HGMamba \cite{cui2025hgmamba}, and Hamba \cite{dong2024hamba}. Unlike these methods, where GCN and Mamba are implemented as independent components, we modify the internal convolutional architecture of the Mamba model and incorporate both forward and backward graph convolutional network. GEM effectively captures both global and local dependencies among joints and demonstrates superior performance compared to alternative approaches. Multiple GEM and Vision Mamba blocks are stacked to form the spatiotemporal refinement module. Furthermore, we propose a bone-joint fusion embedding mechanism to embed and integrate the features of 2D joints and bone vectors. The overall architecture of our framework is illustrated in Figure \ref{fig:backbone1}.

We conducted extensive experiments on the Human3.6M \cite{ionescu2013human3} and MPI-INF-3DHP \cite{mehta2017monocular} datasets to validate the effectiveness of our approach. Our method outperforms most previous methods in terms of both accuracy and computational efficiency. For instance, in comparison with prior sequence-to-frame and sequence-to-sequence architectures, our method achieves higher accuracy while consuming only 1/53 of the computational resources required by PoseFormer \cite{zheng20213d} and 1/9 of those required by MixSTE \cite{zhang2022mixste}. Furthermore, compared to PoseMamba \cite{huang2024posemamba}, our method yields a 0.9 mm reduction in error.

In conclusion, our contributions can be summarized as follows:
\begin{itemize}
\item{We propose a bone aware module that generates the positional information of bone vectors in spherical coordinates, serving as topological guidance to assist the Mamba model in capturing the structural relationships among joint sequences.}
\item{We propose a spatiotemporal refinement module for accurate 3D pose estimation. Additionally, we enhance the convolutional architecture within the Mamba model by integrating forward and backward GCN, which facilitates the capture of local topological dependencies among skeletal joints.}
\item{Extensive experiments on the Human3.6M and MPI-INF-3DHP datasets demonstrate that Mamba-Driven Topology Fusion framework significantly reduces computational complexity and improves accuracy compared to other models.}
\end{itemize}

\section{Related Work}

\subsection{Two-stage 3D Human Pose Estimation}
Two-stage 3D human pose estimators typically extract 2D pose information from images, followed by a 2D to 3D booster. Most methods focus on the design of the booster. Martinez et al. \cite{martinez2017simple} used a multi-layer fully connected neural network to elevate 2D coordinates to 3D. Fang et al. \cite{fang2018learning} enhanced high-level constraints on human pose in 2D to 3D lifting using pose grammar. Wandt et al. \cite{wandt2019repnet} trained a 3D to 2D reprojection network using adversarial training. Li et al. \cite{li2020cascaded} applied data augmentation and designed a deep cascaded fully connected neural network to regress 3D coordinates. Due to the powerful global modeling capability of transformers, they can significantly improve accuracy in 3D HPE tasks. Shan et al. \cite{shan2022p} employed pretraining and fine-tuning to better capture joints dependencies. Zhang et al. \cite{zhang2022mixste} proposed a spatiotemporal alternating network structure to capture joint spatial correlation and temporal motion. Some methods focus on exploring trajectory information across pose time frames. Pavllo et al. \cite{pavllo:videopose3d:2019} used dilated temporal convolutions to explore frame relationships. Li et al. \cite{li2022exploiting} introduced a strided transformer to predict single-frame poses from a sequence. Other methods \cite{zhu2023motionbert, mehraban2024motionagformer, tang20233d, tang2023ftcm, li2022mhformer} also explore temporal information within models to improve accuracy. However, while transformer-based methods enhance model performance, they also reduce execution efficiency. Although some works \cite{einfalt2023uplift, zhao2023poseformerv2, li2024hourglass, zeng2022deciwatch} aim to reduce computational cost, the quadratic computational complexity of the attention mechanism remains a challenge.

\subsection{State Space Model}
Recently, state space model (SSM) have shown great potential in modeling long-range sequences across multiple modalities. The Structured State Space Sequence model (S4) \cite{gu2021efficiently} improves upon SSM by using HiPPO matrices with conditioning matrices \(A\), incorporating discretization techniques, and convolutional representations. Mamba \cite{gu2023mamba} further enhances S4 by introducing parameter-dependent improvements in SSM and employing a parallel scanning algorithm for efficient model computation. Subsequently, Vision Mamba \cite{zhu2024vision} proposed a general-purpose visual backbone with bidirectional Mamba blocks, and VMamba \cite{liu2024vmamba} utilized 2D Selective Scan (SS2D) along four scanning paths to help understand the non-sequential structure of 2D visual data. DVMSR \cite{lei2024dvmsr} employed Vision Mamba and a distillation strategy to achieve state-of-the-art (SOTA) performance in image super-resolution tasks. PoseMamba \cite{huang2024posemamba} and PoseMagic \cite{zhang2024pose} employ the Mamba model for 3D HPE, achieving reduced computational cost while maintaining favorable performance. Motivated by the strong performance of Mamba, we further explore its topological capabilities in the context of 3D HPE.

\subsection{Human Topology Analysis in 3D HPE}
In 3D HPE, analyzing the human topology and processing different structures to reduce errors is a common approach. Xu et al. \cite{xu2020deep} decoupled the 3D pose into bone lengths and bone orientations, predicting them separately using a two-stream structure. Cai et al. \cite{cai2024disentangled} employed a diffusion model with Transformers for denoising noisy bone lengths and orientations, targeting parent and child joints. Xue et al. \cite{xue2022boosting} divided the human body into five parts, exploring their temporal dependencies and relationships with other actions. Furthermore, Graph Convolutional Networks (GCN) effectively represent the topological relationships between human joints using adjacency matrices. 
Yin et al. \cite{yin2023multibranch} utilized multi-branch attention graph convolution to extract feature information contributing to different nodes. Cai et al. \cite{cai2019exploiting} utilized GCN-based methods to integrate domain knowledge of human body layout. While GCN is a lightweight model, it often lacks high accuracy. Wang et al. \cite{wang2019not} proposed an advanced method for handling joints with different degrees of freedom, but it cannot impose strong constraints on all joints. Wu et al. \cite{wu2021limb} applied relative and absolute bone angle constraints on limbs; however, this limitation is flawed as identical relative and absolute bone angles may occur in different actions. Inspired by these methods, we propose using bone vectors in spherical coordinates as global topological information for the human body, where the relationship between two joints is both intuitive and unique.

\section{Method}
\begin{figure*}[ht]
	\centering
	\includegraphics[width=\linewidth]{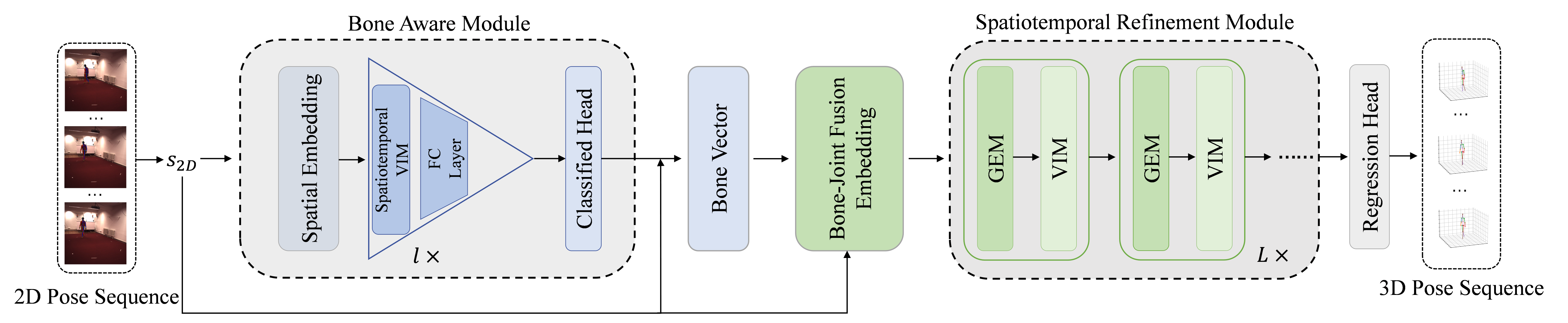}
	\caption{The overall structure of the Mamba-Driven Topology Fusion framework. We perform seq2seq estimation using 2D pose sequences as input. \( s_{2D} \) represents the 2D pose sequence detected from images. First, the bone aware module predicts the category of bone vectors based on \(s_{2D}\), and the corresponding bone vector coordinates are then computed by combining the predicted categories with the 2D joint positions. Then, the bone vectors and 2D poses are combined and fed into the bone-joint fusion embedding. Finally, the spatiotemporal refinement module regresses accurate 3D pose sequences.}
	\label{fig:backbone1}
\end{figure*}
The Mamba-Driven Topology Fusion framework we propose is shown in Figure \ref{fig:backbone1}. Our input is a 2D pose sequence frame \( s_{2D}\) obtained from the 2D pose detector. We first train the bone aware module to compute the polar angle categories for each bone in the spherical coordinate system. Then, we assign values to these categories and infer the estimated spherical coordinates \( bone_s\in\mathbb{R}^{f\times j \times 3} \) for each bone based on the human body structure, where the last dimension corresponds to \( r \), representing the radial distance, \( \theta \), representing the polar angle, and \( \phi \), representing the azimuthal angle. Next, we feed \( s_{2D} \) and \( bone_s \) into the bone-joint fusion embedding for feature fusion, followed by the spatiotemporal refinement module to capture both spatial and temporal information of the joints. Finally, a regression head is used to predict the true 3D pose \( s_{3D}\in\mathbb{R}^{f\times j \times 3} \). In the following section, we first present the preliminaries of SSM, followed by a description of the Mamba-based GEM block, and then detail the proposed components.

\subsection{Preliminary}
\label{Preliminary}
Recent works have demonstrated the strong modeling capabilities of state space model (SSM) \cite{gu2021combining} for long sequence data. SSM assume that a dynamic system can be predicted based on its state at time \( t \). It maps a one-dimensional sequence \( x(t) \in\mathbb{R}^{L} \) to \( y(t) \in\mathbb{R}^{L} \) through a compressed state representation \( h(t) \in\mathbb{R}^{N} \), which encodes the information from the previously scanned sequence. The SSM can be formulated as a linear ordinary differential equation (ODE):
\begin{equation}
	\begin{aligned}h^{\prime}(t)&=Ah(t)+Bx(t)\\
					y(t)&=Ch(t)\end{aligned}
\end{equation}
where \(A \in\mathbb{R}^{N\times N}\), \(B \in\mathbb{R}^{N\times 1}\),and \(C \in\mathbb{R}^{1\times N}\) represent the parameters of SSM. S4 \cite{gu2021efficiently} uses Zero-Order Hold to solve the ODE equation with a step size \( \Delta \). The solution to the discretized ODE equation using \( x_k\in\mathbb{R}^{L\times D} \) can be expressed as:
\begin{equation}
	\label{func:ssm}
	\begin{aligned}
	h_k &= \bar{A}h_{k-1} + \bar{B}x_k\\
	y_k &= Ch_k.
\end{aligned}
\end{equation}
To address the limitations of linear time-invariant SSM, Mamba \cite{zhu2024vision} introduces a new input-dependent parameterization and a hardware-aware parallel algorithm that enables recurrent computation through scanning. A more detailed introduction of Mamba is provided in \cite{gu2023mamba}.

\subsection{GCN-Enhanced Vision Mamba}
\label{section: VMB and GMB}
\begin{figure*}[ht]
	\centering
	\includegraphics[width=\linewidth]{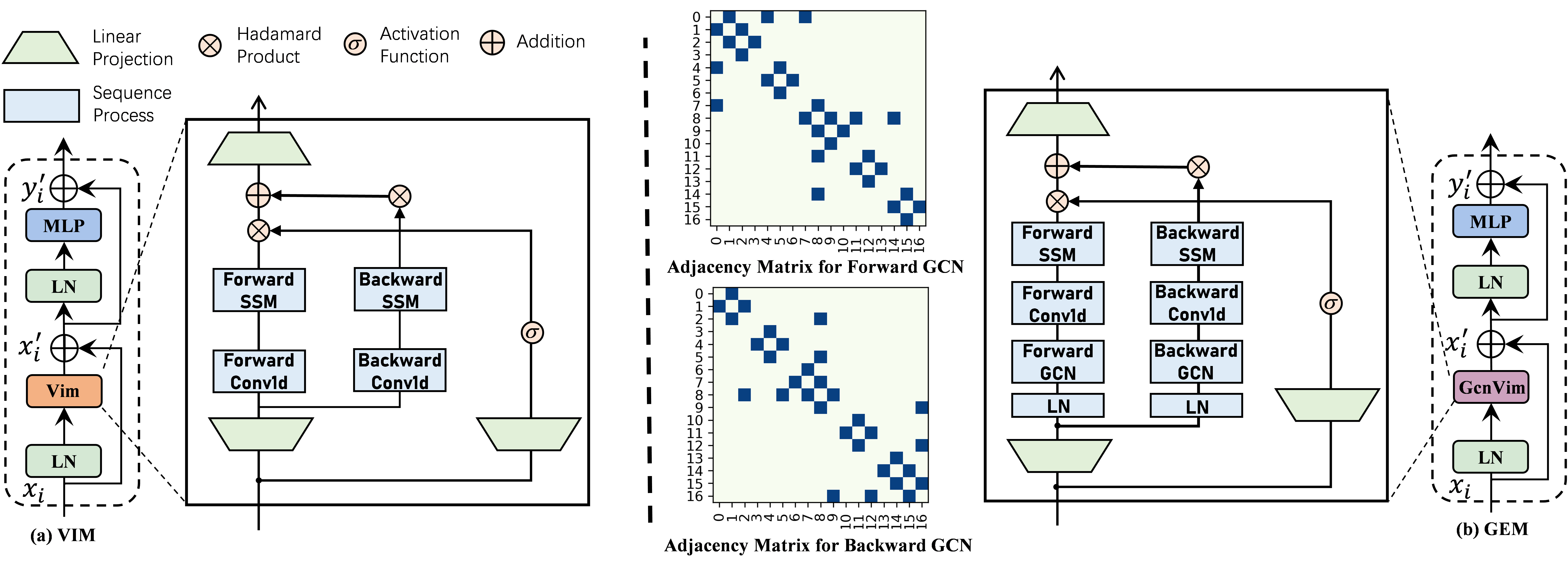}
	\caption{The detailed components of the VIM and GEM blocks are illustrated. Subfigure a illustrates the overall architecture of VIM along with its internal \(Vim\) module. Subfigure b presents the structure of GEM and details the convolutional components enhanced with graph convolution.}
	\label{fig:vmbgmb}
\end{figure*} 

Vim in \cite{zhu2024vision} adapts the Mamba framework for visual tasks by designing a bidirectional SSM that scans the token sequence in both forward and backward directions. The architecture is illustrated in Figure \ref{fig:vmbgmb}-a. Given an input sequence \( x \in \mathbb{R}^{b \times l \times d} \), where \( b \) is the batch size, \( l \) is the sequence length, and \( d \) is the hidden state dimension, the input is first processed by two independent linear projections to increase the state dimensions \(d\) to the expanded state dimensions \(e\), resulting in \( \hat{x} \in \mathbb{R}^{b \times l \times e} \) and \( \hat{z} \in \mathbb{R}^{b \times l \times e} \).
\begin{equation}
		\hat{x} = xW_x, \hat{z} = xW_z
\end{equation}
where \(W_x\) and \(W_z\)  \(\in \mathbb{R}^{d \times e}\). 
The forward and backward passes follow a similar procedure by using an SSM module equipped with convolution:
\begin{equation}
	\begin{aligned}
		\tilde{x}_f &= SSM_f(\sigma(Conv1d_f(\hat{x}))) \\
		\tilde{x}_b &= SSM_b(\sigma(Conv1d_b(flip(\hat{x}))))
	\end{aligned}
\end{equation}
where \(f\) and \(b\) represent the forward and backward processes, \(\tilde{x} \in \mathbb{R}^{b \times l \times e}\), \(\sigma (\cdot) \) is the SiLU activation function, \(Conv1d(\cdot)\) is the causal 1D convolution, and \(SSM(\cdot)\) represents the computation process in function \ref{func:ssm}.
Then, the forward output \( \tilde{x}_f \) and the backward output \( \tilde{x}_b \) are gated by \( \hat{z} \):
\begin{equation}
	\label{funcy_fy_b}
	\begin{aligned}
		y_f &= \tilde{x}_f \circ SiLU(\hat{z})\\
		y_b &= flip(\tilde{x}_b) \circ SiLU(\hat{z})
	\end{aligned}
\end{equation}
where \(\circ\) represents a Hadamard Product.
Finally, The gated outputs are summed and the sequence feature dimension is projected back to the hidden feature dimension \(d\), with a residual connection:
\begin{equation}
	\label{functildey}
	\tilde{y} = x+(y_f+y_b)W_m
\end{equation}
where \(W_m \in \mathbb{R}^{e \times d}\), \(\tilde{y} \in \mathbb{R}^{b \times l \times d}\).

The causal 1D convolution in mamba is useful for processing time series. However, this formulation overlooks the critical need to explicitly capture local joint interdependencies inherent in human kinematics. The relationship between human joints exhibits a physiological topological structure. For instance, as visualized in Figure \ref{fig:boneaware}, the left foot (joint 6) demonstrates strong biomechanical coupling with the left knee (joint 5), but significantly weaker correlations with the contralateral right foot (joint 3). To bridge this gap, we propose integrating Graph Convolutional Networks (GCN) into the convolution process. GCN excel at encoding such anatomical relationships through a joint adjacency matrix, where edge weights reflect anatomical connectivity.

To further enhance bidirectional modeling of joint interactions, we introduce a dual GCN architecture comprising both forward and backward graph propagation streams in Vision Mamba. This design contrasts with decoupled frameworks in prior works like PoseMagic \cite{zhang2024pose} and Hamba \cite{dong2024hamba}, where independent GCN and Mamba may result in the loss of local structural details captured by GCN. By seamlessly integrating GCN prior to the 1D convolutional layers, our approach ensures that subsequent casual convolutions operate on features pre-refined with anatomical context.

Specifically, we add a GCN structure before the 1D convolution in the SSM to process the expanded state sequence \( \hat{x} \). As shown in Figure \ref{fig:vmbgmb}-b, for the normalized and projected \( \hat{x} \):
\begin{equation}
	\bar{x}_f = SSM_f(\sigma(Conv1d_f(GCN_f(LN(\hat{x})))))
\end{equation}
where \(\bar{x}_f \in \mathbb{R}^{b \times l \times e} \), and \(LN(\cdot)\) denotes the layer normalization. As the internal structure becomes more complex, applying layer normalization to \(\hat{x}\) facilitates faster convergence of the model. The \(GCN(\cdot)\) \cite{luo2022learning} is defined as:
\begin{equation}
	GCN_f(\mathring{x}) = \sigma(\mathring{x} + BN( \tilde{D}^{-\frac{1}{2}} \tilde{A} \tilde{D}^{-\frac{1}{2}} \mathring{x} W_1 + \mathring{x}W_2))
\end{equation}
where \( \tilde{A} = A + I_N \) represents the sum of the joint adjacency matrix and the identity matrix, \( \tilde{D}_{ii} = \sum_{j} \tilde{A}_{ij} \), \( W_1 \) and \( W_2 \) are trainable parameter matrices, \( BN(\cdot) \) denotes batch normalization, and \( \sigma(\cdot) \) is the ReLU activation function.
To match the bidirectional SSM in Vision Mamba, we set up backward GCN to process the joint adjacency structure of the reverse sequence:
\begin{equation}
	\bar{x}_b = SSM_b(\sigma(Conv1d_b(GCN_b(LN(flip(\hat{x}))))))
\end{equation}
where \( GCN_b(\cdot)\) has the same structure as \( GCN_f(\cdot) \), except that the adjacency matrix \( A \) in \( \tilde{A} \) is modified according to the adjacency relationships of the reverse sequence. Since there exists bidirectional connectivity between adjacent joints, the forward and backward adjacency matrices \(A\) are symmetric. Their structures are illustrated in Figure \ref{fig:vmbgmb}-b. The subsequent computational process remains consistent with Equations \ref{funcy_fy_b} and \ref{functildey}.

Finally, to further enhance the model's expressive power, we embed Vim and GCN-enhanced Vim into a transformer-like encoder structure. As shown in Figure \ref{fig:vmbgmb} a and b, given the input pose feature sequence \(x_i\) at the \(i\)-th layer, it first passes through the Vim or GCN-enhanced Vim, followed by a MLP with a residual connection. We denote these two methods as \( VIM(\cdot) \) and \( GEM(\cdot) \) as follows:
\begin{equation}
\label{vim}
	\begin{aligned}
	x^{\prime}_i &= Vim/GcnVim(LN(x_i)) + x_i \\
	y^{\prime}_i &= MLP(LN(x^{\prime}_i)) + x^{\prime}_i
	\end{aligned}
\end{equation}
The following module adopts these two blocks as fundamental building components of the model.

\subsection{Bone Aware Module}
\begin{figure}[ht]
	\centering
	\includegraphics[width=\linewidth]{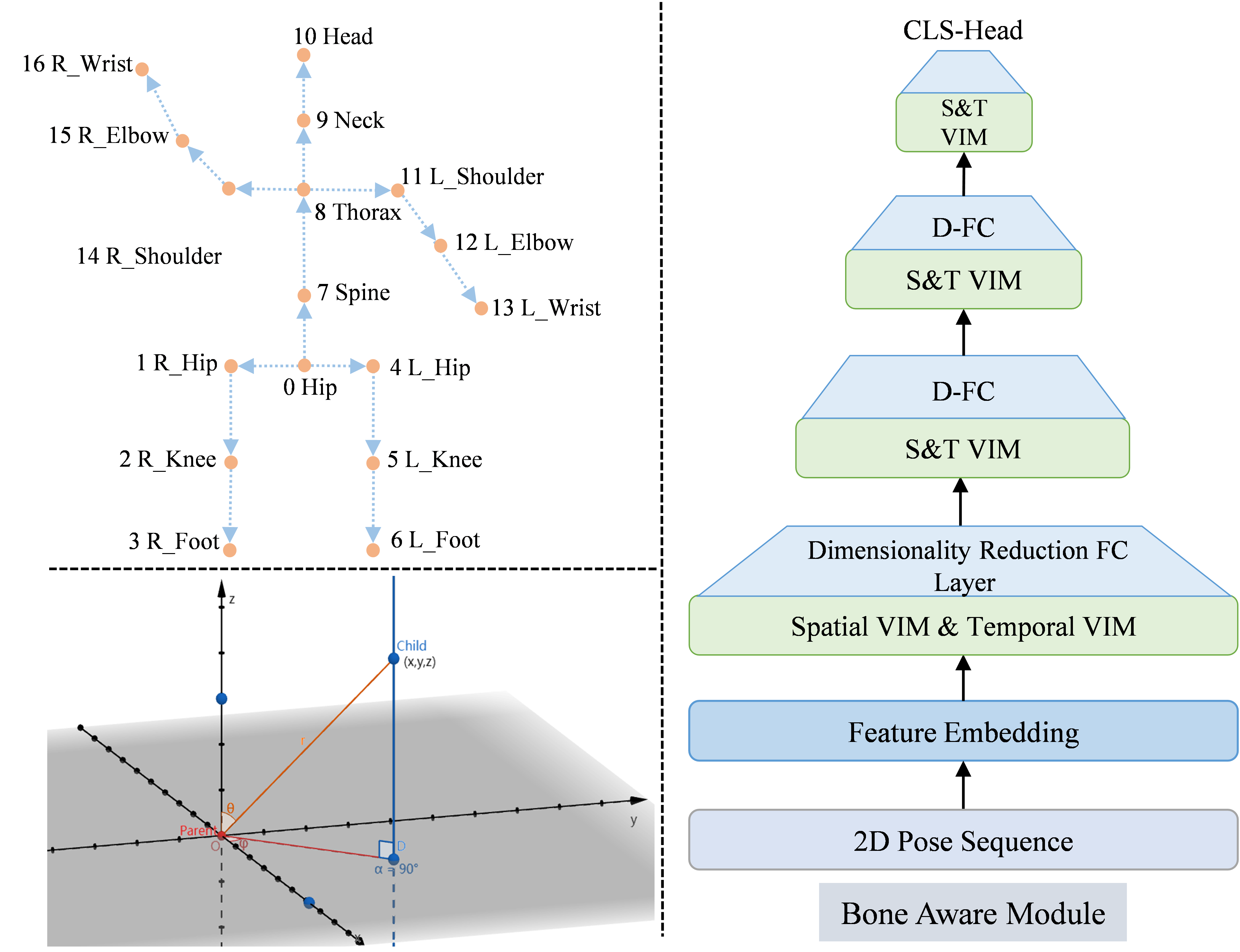}
	\caption{The top-left subfigure shows the topology of the human skeleton. The start and end points of each arrow represent the parent and child joints of the bone, respectively. The bottom-left subfigure represents the placement of a skeleton in a spherical coordinate system, where the parent joint is aligned with the coordinate origin \(O\). The right-side subfigure illustrates the model structure of our proposed bone aware module. D-FC stands for Dimensionality Reduction Fully Connected layer, and S\&T VIM represents spatial and temporal VIM processing.}
	\label{fig:boneaware}
\end{figure}

The bone aware module computes the direction and length of bone vectors, providing human skeletal topology as guidance for Mamba during joint position inference. This partially addresses Mamba's limited capacity to capture human body topology when processing the entire pose sequence. The bone vector in 3D space contains depth positioning information of the child joint relative to the parent joint. As shown in Figure \ref{fig:boneaware}, a bone vector in space is defined as \(l_i = p^{child}_i - p^{parent}_i\), where \(p^{child}_i\) and \(p^{parent}_i\) represent the child and parent joints of the \(i\)-th bone, respectively. This vector is then transferred to the spherical coordinate system, where the parent joint \(p^{parent}_i\) is aligned with the origin \(o\) of the spherical coordinate system, and the child joint can be represented by spherical coordinates \((r_i, \theta_i, \phi_i)\). If the spherical coordinates of the child joint for each bone vector can be estimated, the direction and length of the bone vector can be roughly determined, which helps guide the model to capture the correct human body topology.

First, the conversion of bone vectors between spherical and Cartesian coordinate systems is analyzed. Let \(x_i, y_i, z_i\) represent the coordinates of the \(i\)-th bone vector in the Cartesian coordinate system, then the conversion is as follows:
\begin{equation}
\label{eq1}
\begin{aligned}
	r_i &= \sqrt{x_i^2 + y_i^2 + z_i^2} \\
	\theta_i &= \arctan\left(\frac{\sqrt{x_i^2 + y_i^2}}{z_i}\right) \\
	\phi_i &= \arctan\left(\frac{y_i}{x_i}\right)
\end{aligned}
\end{equation}
where \(z_i\) is unknown, \(x_i\) and \(y_i\) can be obtained from the 2D pose information, with \(\theta_i\) ranging from 0 to \(\pi\) and \(\phi_i\) ranging from 0 to \(2\pi\).
In Equation \ref{eq1}, the polar angle \(\theta_i\) contains depth information, and inferring the polar angle allows further calculation of the radial distance \(r_i\). However, directly regressing the polar angle is challenging, so we perform classification to predict it. By dividing the range of the polar angle into categories of size \(\pi/n\), we can classify it into \(n\) categories, each representing an approximate direction of the bone's depth. 

Next, we design the bone aware module to predict the polar angle category of each bone vector in the human skeleton. It takes a 2D pose sequence \(s_{2D}\) as input, followed by a pyramid-shaped classification network with feature attenuation. As shown in Figure \ref{fig:boneaware}, given \( s_{2D}\), we first apply a feature embedding to map it to a high-dimensional feature \( s^{\prime}_{2D} \in \mathbb{R}^{f \times j \times d} \). The network of the bone aware module consists of \(l\) layers, each containing a spatial VIM, a temporal VIM, and a feature decay block. For the computation at the \(i\)-th layer, we first use VIM to capture the spatial features of the bone structure:
\begin{equation}
	s^{s}_i = VIM(s^{\prime}_i)
\end{equation}
where \(s^{s}_i \in \mathbb{R}^{f \times j \times d}\), \(i \in \{0,...,l\}\), and \(s^{\prime}_0=s^{\prime}_{2D}\). Then, the output of the spatial VIM is transposed and fed into the temporal VIM to capture the temporal trajectory features of the bones:
\begin{equation}
	s^{t}_i = VIM(trans(s^{s}_i))
\end{equation}
where \(s^{t}_i \in \mathbb{R}^{j \times f \times d}\), and \( trans \) represents the transpose operation applied to the temporal and spatial dimensions. To reduce the computational load and compress the model features, we apply a feature attenuation matrix to \(s^{t}_i\):
\begin{equation}
	s^{\prime}_{i+1}=trans(s^{t}_i)W^i_R
\end{equation}
where \( W^i_R \in \mathbb{R}^{(d\cdot {\frac{1}{2}}^{i}) \times (d\cdot {\frac{1}{2}}^{i+1})} \), and \( s^{\prime}_{i+1} \in \mathbb{R}^{f \times j \times (d\cdot {\frac{1}{2}}^{i+1}) }\). The feature attenuation ratio in each layer is set to 2, except for the final layer, where we replace feature attenuation with a classification head to classify the polar angle category of each skeletal joint at each timestamp: 
\begin{equation}
	c = cls\text{-}head(trans(s^{t}_l))
\end{equation}
where \(c \in \mathbb{R}^{f\times j\times n}\), \(cls\text{-}head\) is a linear projection followed by a softmax function. The classification result can be obtained by returning the index of the maximum value in the softmax output.

Finally, we assign values to the classification results of the bone aware module, replacing each category with the midpoint of its corresponding angular range. For example, \(\frac{1}{2n}\pi\) is used to represent the first category. The assigned result \(\theta \in \mathbb{R}^{f \times j \times 1}\) can roughly indicate the direction of the polar angle. Given the \(x\) and \(y\) coordinates from the 2D pose, \(z\) can be inferred using the following equation:
\begin{equation}
	z=\frac{\sqrt{x^2+y^2}}{tan\theta}.
\end{equation}
Through Equation \ref{eq1}, the bone vectors in the spherical coordinate system \( B_{3D} = (r, \theta, \phi) \) can be further computed, where \( B_{3D} \in \mathbb{R}^{f \times j \times 3} \). The resulting \( B_{3D}\) encodes both the length and direction of bones, representing an approximate relative position and distance between child and parent joints. Although it does not provide exact positional information, it serves as a coarse structural prior that effectively guides the subsequent network in localizing joint positions.

\subsection{Bone-Joint Fusion Embedding}
\begin{figure}[ht]
	\centering
	\includegraphics[width=\linewidth]{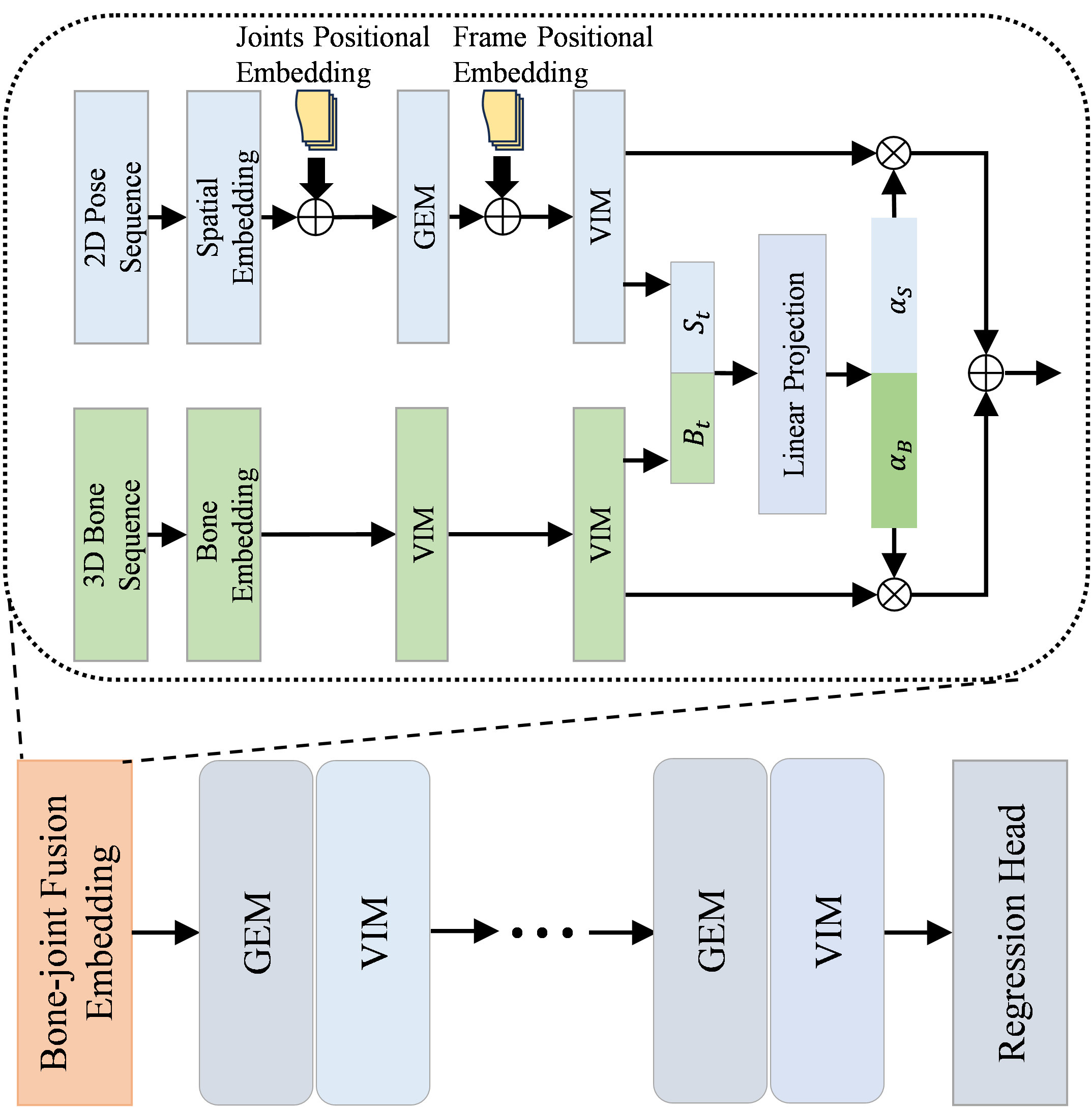}
	\caption{The structural diagrams of the spatiotemporal refinement module and the bone-joint fusion embedding. The bone-joint fusion embedding serves as the head of the spatiotemporal refinement module, processing the input 2D pose sequences and 3D bone sequences.}
	\label{fig:bonejointfusionhead}
\end{figure}
Before feeding the 2D pose and bone vectors into the refinement network, it is necessary to fuse information from the two representations. As they reside in different coordinate systems, we first map both representations into high-dimensional feature spaces, extract salient features, and then perform feature fusion. The structure of the bone-joint fusion embedding is illustrated in Figure \ref{fig:bonejointfusionhead}. For the 2D pose sequence branch, the 2D sequence is first mapped to high-dimensional embedding features, followed by adding joints positional embedding \(E_s \in \mathbb{R}^{j \times D}\), and then processed by a GEM block:
\begin{equation}
	S_s = GEM(s_{2D}M_s+E_s).
\end{equation}
Then, along the time dimension, add a frame positional embedding \(E_t \in \mathbb{R}^{f \times D}\) to \(S_s\) and feed it into the VIM block:
\begin{equation}
	S_t = VIM(trans(S_s)+E_t).
\end{equation}
For the bone vector branch, we use the same processing approach as the 2D pose branch, but without adding positional embedding. Instead, we use VIM to process spatial relationships, as the graph convolutional structure in GEM is designed for joints. This can be expressed as follows:
\begin{equation}
	B_t = VIM(trans(VIM(B_{3D}M_b))).
\end{equation}
Since 2D poses and bone vectors are two different attributes, the features processed through spatiotemporal extraction are more conducive to the fusion of these two types of information. Ultimately, we adopt an adaptive fusion method \cite{zhu2023motionbert} to aggregate both sequence information, which is defined as:
\begin{equation}
	X_0 = \alpha_S \otimes S_t + \alpha_B \otimes B_t
\end{equation}
where \(\otimes\) represents element-wise multiplication, and \(X_0\) is the fused feature. The adaptive fusion weights \(\alpha_S\) and \(\alpha_B\) are defined as:
\begin{equation}
	\alpha_S, \alpha_B = softmax(W \cdot Concat(S_t, B_t))
\end{equation}
where \(W\) is a learnable weight matrix.

\subsection{Spatiotemporal Refinement Module}
To predict the final 3D pose sequence \(s_{3D}\), \(X_0\) is sent into the spatiotemporal refinement module. As shown in Figure \ref{fig:bonejointfusionhead}, the spatiotemporal refinement module consists of multiple layers of basic blocks, each layer comprising a series of GEM and VIM blocks. The computation process for the \(i\)-th layer can be expressed as:
\begin{equation}
	X_{i+1} = trans(VIM(trans(GEM(X_i))))
\end{equation}
where \(i\in \{0,1,\dots, L\}\), \(X_{i+1}\in \mathbb{R}^{f\times j\times D}\).
GEM is used to handle the relationships between local and global keypoints in the spatial pose sequence, while VIM is used to capture the relationships of the same keypoint across consecutive time frames in the temporal pose sequence. After multiple layers of spatiotemporal refinement, a regression head is used to compute the precise 3D pose sequence \( s_{3D} \in \mathbb{R}^{f \times j \times 3} \).

\section{Experiments}

\subsection{Dataset and evaluation metric}

We evaluate our method on two 3D HPE bench-mark datasets: Human3.6M \cite{ionescu2013human3} and MPI-INF-3DHP \cite{mehta2017monocular}. 
\subsubsection{Human3.6M} The \text{Human3.6M} dataset consists of 3.6 million video frames captured in indoor settings, featuring 11 professional actors performing 15 different actions across 4 synchronized camera views. For training and testing, we follow the approach in \cite{zhao2023poseformerv2, pavllo:videopose3d:2019}, using data from 5 subjects (S1, S5, S6, S7, S8) for training and 2 subjects (S9, S11) for testing. We evaluate model performance using two protocols, as in prior work \cite{zhang2022mixste, pavllo:videopose3d:2019}. Protocol 1 calculates the mean per-joint position error (MPJPE), which measures the average Euclidean distance in millimeters between predicted and ground-truth 3D keypoints. Protocol 2 uses procrustes MPJPE (P-MPJPE), which computes the reconstruction error after aligning the predicted 3D pose to the ground-truth using procrustes analysis. Moreover, following Pavllo \textit{et al.} \cite{pavllo:videopose3d:2019}, we evaluate the model's MPJVE, which is the mean Euclidean distance of the first derivative.

\subsubsection{MPI-INF-3DHP} The \text{MPI-INF-3DHP} dataset, another popular benchmark for 3D human pose estimation, includes 1.3 million frames from both controlled indoor environments and complex outdoor scenarios. Following the setup in \cite{zhang2022mixste, mehraban2024motionagformer}, we use MPJPE, the percentage of correct keypoints (PCK) with a 150mm threshold, and the area under the curve (AUC) as evaluation metrics.

\subsection{Implementation Details}
The experiments are conducted using PyTorch and run on an RTX 4090 GPU. 2D pose sequences are obtained by the Stacked Hourglass (SH) networks \cite{2016Stacked}. The training process is divided into two stages. In the first stage, the bone aware module is trained using a cross-entropy loss function to optimize the model. After the training of the first stage is completed, the parameters of the bone perception module are frozen. In the second stage, only the spatiotemporal refinement module is trained, using weighted MPJPE, N-MPJPE, and MPJVE as the loss functions, with weights of 1, 0.5, and 20, respectively. The learning rate, learning rate decay, and dropout for the first stage are set to \(2 \times 10^{-3}\), \(0.99\), and \(0.1\), respectively. For the second stage, the learning rate, learning rate decay, and dropout are set to \(5 \times 10^{-4}\), \(0.99\), and \(0.1\), respectively. Both modules are trained using the AdamW optimizer \cite{loshchilov2017decoupled} with weight decay of \(0.01\). The bone aware module is trained for 60 epochs with a batch size of 128, and the spatiotemporal refinement module is trained for 120 epochs with a batch size of 16.

In the SSM part of VIM and GEM, we set the hidden state dimension \(d\) equal to the expanded state dimension \(e\). The depth and hidden feature dimension of the bone aware module are set to 4 and 64, respectively. The number of classification categories for bone polar angles, \(n\), is set to 6. For the spatiotemporal refinement module, we define two versions: Ours-tiny and Ours-large. The depth and hidden feature dimensions for Ours-tiny and Ours-large are set to \(\{L=8, D=64\}\) and \(\{L=12, D=128\}\), respectively.

\subsection{Quantitative Evaluation}

\begin{table}[ht]\Large
	\caption{Comparison of errors and computational costs between our method and previous transformer-based SOTA approaches. \(T\) represents the number of frames actually used by the model. (Top) sequence-to-frame methodology, (Middle) sequence-to-sequence methodology.}
	\centering
	\resizebox{\linewidth}{!}{
		\begin{tabular}{l|cccc}
			\toprule
			Method                                          	& $T$ & Param  & MACs/frame & MPJPE$\downarrow$	 \\
			\midrule
			Poseformer~\cite{zheng20213d}ICCV'2021				& 81  & 9.6M  & 679M	&44.3mm\\ 
			Xue~\textit{et al.}~\cite{xue2022boosting}TIP'22	& 243 & 6.2M  & 1.37G	&43.1mm\\
			Li~\textit{et al.}~\cite{li2022exploiting}TMM'23	& 351 & 4.3M  & 1.07G &43.7mm\\
			PoseformerV2~\cite{zhao2023poseformerv2}CVPR'23		& 243 &14.3M  & 528M	&45.2mm\\
			\midrule
			MixSTE~\cite{zhang2022mixste}CVPR'22						& 243 & 33.8M & 572M	&40.9mm\\
			STCFormer~\cite{tang20233d}CVPR'23							& 243 & 18.9M & 321M	&40.5mm\\
			MotionAGFormer-S~\cite{mehraban2024motionagformer}WACV'24 	& 81 & 4.8M  & 81M	    &42.5mm\\
			Zhong~\textit{et al.}~\cite{zhong2023frame}TMM'24			& 243 & 42.3M & 418M	&40.1mm \\
			PoseMamba~\cite{huang2024posemamba}AAAI'2025				& 243 & 3.4M &57M & 40.8mm\\
			HGMamba~\cite{cui2025hgmamba}IJCNN'2025						& 81  & 6.1M &99M & 42.8mm \\
			\midrule
			Ours-tiny											& 243 & 0.9M  & 12.8M	&41.7mm\\
			Ours-large								& 243 & 4.4M  & 60.7M	&40.0mm\\
			
			%			Ours-tiny											& 351 & 0.9M  & 17.9M	&\textbf{40.8mm}\\
			\bottomrule
		\end{tabular}%
	}
	\label{tab:compare}
\end{table}

\begin{table*}[ht]\Large
	\caption{Quantitative comparisons for each action on Human3.6M using detected 2D pose sequence and 2D ground truth.\(T\) represents the input time frame. (*) denotes using HRNet \cite{sun2019deep}. \(\dag\) indicates the use of 2D ground truth as inputs. The best results are highlighted in bold, and the second-best results are underlined.}
	\centering
	\resizebox{\linewidth}{!}{
		\begin{tabular}{lc|ccccccccccccccc|c}
			\toprule\toprule
			\textbf{MPJPE} & $T$ & Dire. & Disc. & Eat & Greet & Phone & Photo & Pose & Purch. & Sit & SitD & Smoke & Wait & WalkD & Walk & WalkT & Avg\\
			\midrule
			Poseformer~\cite{zheng20213d}ICCV'21					& 81 & 41.5 &44.8 &39.8 &42.5 &46.5 &51.6 &42.1 &42.0 &53.3 &60.7 &45.5 &43.3 &46.1 &31.8 &32.2 &44.3\\
			*MHFormer~\cite{li2022mhformer}CVPR'22					& 351 & 39.2 			& 43.1 				& 40.1 & 40.9 			 & 44.9 & 51.2 					& 40.6 & 41.3 & 53.5 & 60.3 & 43.7 & 41.1 & 43.8 & 29.8 & 30.6 & 43.0\\
			MixSTE~\cite{zhang2022mixste}CVPR'22 	& 243 &\textbf{37.6} &\underline{40.9} &\textbf{37.3}& 39.7 &\underline{42.3} &\underline{49.9} &40.1 &39.8 & 51.7 &\underline{55.0} & 42.1 & 39.8 & 41.0 &27.9 &\underline{27.9} & 40.9\\
			P-STMO~\cite{shan2022p}ECCV'22 							& 243 &38.9 & 42.7 				& 40.4 & 41.1 			 & 45.6 & \textbf{49.7} 		& 40.9 & 39.9 & 55.5 & 59.4 & 44.9 & 42.2 & 42.7 & 29.4 & 29.4 & 42.8\\
			StridedFormer~\cite{li2022exploiting}TMM'23 			& 351 & 40.3 			& 43.3 				& 40.2 & 42.3 			 & 45.6 & 52.3 					& 41.8 & 40.5 & 55.9 & 60.6 & 44.2 & 43.0 & 44.2 & 30.0 & 30.2 & 43.7\\
			Einfalt~\textit{et al.}~\cite{einfalt2023uplift}WACV'23 & 351 & 39.6 			& 43.8 				& 40.2 & 42.4 			 & 46.5 & 53.9 					& 42.3 & 42.5 & 55.7 & 62.3 & 45.1 & 43.0 & 44.7 & 30.1 & 30.8 & 44.2\\
			STCFormer~\cite{tang20233d}CVPR'23 						& 243 & 39.6 			& 41.6 				&\underline{37.4} & 38.8 & 43.1 & 51.1 					& \underline{39.1} & 39.7 & \underline{51.4} & 57.4 &\textbf{41.8} &\underline{38.5} & 40.7 &\textbf{27.1} & 28.6 & 41.0\\
			FTCM~\cite{tang2023ftcm}TCSVT'24						& 351 & 42.2 			& 44.4 				& 42.4 & 42.4 			 & 47.7 & 55.8 					& 42.7 & 41.9 & 58.7 & 64.5 & 46.1 & 44.2 & 45.2 & 30.6 & 31.1 & 45.3\\
			PoseMamba~\cite{huang2024posemamba}AAAI'2025    &243 &\underline{38.8} &\textbf{40.8} &38.8 &\underline{35.2} &\textbf{42.1} &50.8 &\textbf{38.8} &\textbf{36.4} &51.8 &61.9 &\underline{42.0} &\textbf{38.4} &\underline{38.7} &28.1 &28.7 &\underline{40.8} \\
			\midrule
			Ours-tiny								& 243 & 39.9 			& 43.2 				& 38.2 &36.1 & 44.6 & 52.4 					& 40.9 & \underline{38.1} & 54.8 & 57.8 & 43.3 & 39.9 &39.3 & 28.6 & 29.0 & 41.7\\
%			Ours-tiny								& 351 & \underline{38.4} 			& 42.6 				& 37.8 &\underline{35.3} & 43.0 & 51.9 					& \textbf{38.5} & 38.7 & 51.6 & 57.0 & 42.7 & 39.7 &\underline{39.3} & 27.6 & 28.6 & \underline{40.8}\\
			Ours-large		& 243 & 39.4 &41.3 	& 38.0 &\textbf{34.6} &42.5 &51.7 &\underline{39.1} &\textbf{36.4} &\textbf{51.3} &\textbf{54.5} &\textbf{41.8} &38.9 &\textbf{37.0} & \underline{27.4} &\textbf{27.6} & \textbf{40.0}\\
			\midrule\midrule
			
			MixSTE~\cite{zhang2022mixste}CVPR'22 \dag		  & 243 &21.6  &22.0  &20.4  &21.0  &20.8  &24.3  &24.7  &21.9  &26.9  &24.9  &21.2  &21.5  &20.8  &14.7  &15.7  &21.6 \\
			Xue~\textit{et al.}~\cite{xue2022boosting}TIP'22 \dag  & 243 &25.8  &25.2  &23.3  &23.5  &24.0  &27.4  &27.9  &24.4  &29.3  &30.1  &24.9  &24.1  &23.3  &18.6  &19.7  &24.7 \\
			StridedFormer~\cite{li2022exploiting}TMM'23\dag       & 243 &27.1  &29.4  &26.5  &27.1  &28.6  &33.0  &30.7  &26.8  &38.2  &34.7  &29.1  &29.8  &26.8  &19.1  &19.8  &28.5 \\
			Zhong~\textit{et al.}~\cite{zhong2023frame}TMM'24\dag & 243 &20.6  &19.9  &19.5  &19.3  &19.0  &\underline{21.7} &21.9  &20.7  &23.5  &24.3  &19.9  &18.5  &19.5  &\underline{13.1} &\underline{13.9} &19.7 \\
			KTPformer~\cite{peng2024ktpformer}CVPR'24\dag		  & 243 &\underline{19.6} &\underline{18.6} &\underline{18.5} &\underline{18.1} &\underline{18.7} &22.1  &\underline{20.8} &\underline{18.3} &\underline{22.8} &\underline{22.4} &\underline{18.8} &\underline{18.1} &\underline{18.4} &13.9  &15.2  &\underline{19.0} \\
			
			\midrule
			Ours-large\dag 										  & 243 &\textbf{16.1} &\textbf{16.0} &\textbf{16.0} &\textbf{14.5} &\textbf{15.6} &\textbf{16.0} &\textbf{15.8} &\textbf{16.1} &\textbf{20.7} &\textbf{19.3} &\textbf{16.6} &\textbf{14.3} &\textbf{13.7} &\textbf{10.7} &\textbf{10.5} &\textbf{15.4}\\
			\midrule\midrule
			\textbf{P-MPJPE} 					  				  	& $T$ & Dire.& Disc. & Eat & Greet & Phone & Photo & Pose & Purch. & Sit & SitD & Smoke & Wait & WalkD & Walk & WalkT & Avg\\
			\midrule
			%			PAA~\cite{xue2022boosting}								& 243 &a31.2 &b34.1 &a31.9 & 33.8 &a33.9 &b39.5 &a31.6 &a30.0 & 45.4 &b48.1 &a35.0 &a31.1 & 33.5 &a22.4 &a23.6 &b33.7\\
			Poseformer~\cite{zheng20213d}ICCV'21					&81 &32.5 &34.8 &32.6 &34.6 &\underline{35.3} &\underline{39.5} &32.1 &32.0 &\textbf{42.8} &48.5 &\textbf{34.8} &32.4 &35.3 &24.5 &26.0 &34.6\\
			*MHFormer~\cite{li2022mhformer}CVPR'22 	  			    & 351 &\underline{31.5} & 34.9 & 32.8 & 33.6 &\underline{35.3} &39.6 &\underline{32.0} & 32.2 &43.5 & 48.7 & 36.4 & 32.6 & 34.3 &23.9 & 25.1 &34.4\\
			P-STMO~\cite{shan2022p}ECCV'22			  				& 243 &\textbf{31.3} & 35.2 & 32.9 & 33.9 & 35.4 &\textbf{39.3} & 32.5 & 31.5 & 44.6 &\underline{48.2} &36.3 & 32.9 & 34.4 &\underline{23.8} &\underline{23.9} &34.4\\
			StridedFormer~\cite{li2022exploiting}TMM'23 		 	& 351 & 32.7 & 35.5 &\underline{32.5} & 35.4 & 35.9 & 41.6 & 33.0 & 31.9 & 45.1 & 50.1 &36.3 & 33.5 & 35.1 & 23.9 & 25.0 & 35.2\\
			Einfalt~\textit{et al.}~\cite{einfalt2023uplift}WACV'23 & 351 & 32.7 & 36.1 & 33.4 & 36.0 & 36.1 & 42.0 & 33.3 & 33.1 & 45.4 & 50.7 & 37.0 & 34.1 & 35.9 & 24.4 & 25.4 & 35.7\\
			FTCM~\cite{tang2023ftcm}TCSVT'24					    & 351 & 31.9 & 35.1 & 34.0 & 34.2 & 36.0 & 42.1 & 32.3 &31.2 & 46.6 & 51.9 & 36.5 & 33.8 & 34.4 & 24.0 & 24.9 & 35.3\\
			PoseMamba~\cite{huang2024posemamba}AAAI'2025  			& 243 &32.3 &\underline{34.0} &33.2 &30.2 &\textbf{34.9} &40.6 &\underline{32.0} &\underline{31.0} &44.5 &53.0 &36.3 &\textbf{31.3} &\underline{33.5} &\underline{23.8} &24.6 &\underline{34.3} \\
			\midrule
			Ours-tiny									    	  	& 243 & 32.3 &34.8 &\textbf{32.3} &\underline{29.8} & 36.9 & 41.5 & 32.8 & 32.3 & 45.8 & 50.9 & 36.7 &32.2 &34.0 & 23.9 & 25.1 & 34.8\\
%			Ours-tiny									    	  	& 351 & \textbf{31.1} &\underline{34.6} &\textbf{31.9} &\underline{29.2} & \underline{35.1} & 41.1 & \textbf{31.0} & 33.2 & 43.6 & 51.3 & \underline{36.2} &\underline{32.1} &\underline{33.7} & \underline{23.3} & 24.4 & \underline{34.1}\\
			Ours-large							& 243 & 32.2 &\textbf{33.6} &32.6 &\textbf{29.1} &\textbf{34.9} & 40.9 &\textbf{31.6} &\textbf{30.9} &\underline{43.2} &\textbf{47.6} &\underline{35.7} &\underline{31.5} &\textbf{31.9} & \textbf{22.9} &\textbf{23.6} &\textbf{33.5}\\
			\bottomrule\bottomrule
		\end{tabular}
	}
	\label{tab:human36mp1}
\end{table*}

\begin{table*}[ht]\Large
	\caption{MPJVE Comparison for Each Action on Human3.6M Using Detected 2D Pose Sequences.}
	\centering
	\resizebox{\linewidth}{!}{
		\begin{tabular}{lc|ccccccccccccccc|c}
			\toprule\toprule
			\textbf{MPJVE} 					  				  	& $T$ & Dire.& Disc. & Eat & Greet & Phone & Photo & Pose & Purch. & Sit & SitD & Smoke & Wait & WalkD & Walk & WalkT & Avg\\
			\midrule
			MixSTE~\cite{zhang2022mixste}CVPR'22				&243 &2.5 &2.7 &1.9 &2.8 &1.9 &2.2 &2.3 &2.6 &1.6 &2.2 &1.9 &2.0 &3.1 &2.6 &2.2 &2.3\\
			MotionBERT~\cite{zhu2023motionbert}ICCV'23			&243 &\underline{1.8} &\underline{2.1} &\underline{1.5} &\underline{2.0} &\underline{1.5} &\underline{1.9} &\underline{1.8} &\underline{2.1} &\underline{1.2} &\underline{1.8} &\underline{1.5} &\underline{1.4} &\underline{2.6} &\textbf{2.0} &\underline{1.7} &\underline{1.8}\\
			Zhong~\textit{et al.}~\cite{zhong2023frame}TMM'24	&243 &2.4 &2.5 &1.9 &2.7 &1.8 &2.2 &2.2 &2.6 &1.5 &2.1 &1.8 &2.0 &3.0 &\underline{2.5} &2.1 &2.2\\
			\midrule
			Ours-large											&243 &\textbf{1.7} &\textbf{1.9} &\textbf{1.4} &\textbf{1.9} &\textbf{1.4} &\textbf{1.8} &\textbf{1.7} &\textbf{2.0} &\textbf{1.1} &\textbf{1.6} &\textbf{1.4} &\textbf{1.3} &\textbf{2.4} &\textbf{2.0} &\textbf{1.6} &\textbf{1.7}\\		
			\bottomrule\bottomrule
		\end{tabular}
	}
	\label{tab:human3.6mpjve}
\end{table*}

\subsubsection{Results on Human3.6M} 
We compared our method with previous SOTA methods on the Human3.6M dataset. For a fair comparison, we adopted the input sequence frame length of 243 frames which is utilized by the majority of methods.

The computational cost and accuracy of our method, which includes the spatiotemporal refinement module and the bone aware module, are compared with previous transformer-based SOTA methods in Table \ref{tab:compare}. Compared to sequence-to-frame models, Ours-tiny model has an extremely small number of parameters and computational cost (about 1/53 of PoseFormer \cite{zheng20213d}), while achieving the highest accuracy (41.7 mm on MPJPE).  Compared to sequence-to-sequence models, Ours-large method achieves a win-win in both computational cost and MPJPE (the computational cost is about 1/9 of MixSTE \cite{zhang2022mixste}, with a 0.9 mm reduction in error). Compared to PoseMamba \cite{huang2024posemamba} and HGMamba \cite{cui2025hgmamba}, which are also based on the Mamba model, our method achieves superior accuracy with similar computational complexity. These results demonstrate the effectiveness of the Mamba module in accelerating computation and show that our topological fusion helps Mamba estimate more accurate results.

As shown in Table \ref{tab:human36mp1}, our method outperforms previous SOTA methods on both MPJPE and P-MPJPE (40.0 mm on Protocol \#1 and 33.5 mm on Protocol \#2). Our method performs better on most actions, especially on challenging actions with significant occlusions, such as sitting and sitting down. This indicates that our proposed topology enhanced mamba can effectively capture both local and global topological structures of the keypoints. To explore the upper bound of model performance, results using real 2D pose inputs are reported in the middle of Table \ref{tab:human36mp1}. Our method achieves the best results on all actions compared to previous methods (15.4 mm on Protocol \#1). This shows that using better 2D pose estimation results as input can amplify the expressive power of our model.

The comparison of speed error results is reported in Table \ref{tab:human3.6mpjve}. Our method performs more smoothly across various actions, achieving the smallest MPJVE of 1.7 mm. This demonstrates the effectiveness of the proposed framework in fitting the temporal trajectories of keypoints, resulting in smoother action trajectories.

\begin{table}[ht]\Large
	\centering
	\caption{Quantitative Results on MPI-INF-3DHP. \(f\) represents the number of input frames, \(\uparrow\) indicates that higher values are better, and \(\downarrow\) indicates that lower values are better. The best results are highlighted in bold, and the second-best results are underlined.}
	\resizebox{\linewidth}{!}{
	\begin{tabular}{lc | ccc}
		\toprule
		Model           										  & $f$ & PCK$\uparrow$ & AUC$\uparrow$ & MPJPE$\downarrow$ \\
		\midrule
		MixSTE~\cite{zhang2022mixste}CVPR'22                      & 27 & 94.4 & 66.5 & 54.9 \\
		P-STMO~\cite{shan2022p}ECCV'22                            & 81 & 97.9 & 75.8 & 32.2 \\
		PoseFormerV2~\cite{zhao2023poseformerv2}CVPR'23           & 81 & 97.9 & 78.8 & 27.8 \\
		STCFormer~\cite{tang20233d}CVPR'23						  & 81 & \underline{98.7} & 83.9 & 23.1 \\
		Zhong~\textit{et al.}~\cite{zhong2023frame}TMM'24		  & 81 & 98.6 & 71.5 & 39.4 \\
		FTCM~\cite{tang2023ftcm}TCSVT'24 						  & 81 & 97.9 & 79.8 & 31.2 \\
		MotionAGFormer~\cite{mehraban2024motionagformer}WACV'24   & 81 & 98.2 & 85.3 & 16.2 \\
		RePOSE~\cite{sun2024repose}ECCV'24   					  & 81 & 98.3 & \underline{86.7} & \underline{15.5} \\
		\midrule
		Ours-tiny												  & 81 & 98.5 & 82.3 & 22.8 \\
		Ours-large												  & 81 & \textbf{99.2} & \textbf{87.2} & \textbf{15.3} \\
		\bottomrule
	\end{tabular}
}
	\label{tab:table_3dhp}
\end{table}

\subsubsection{Results on MPI-INF-3DHP} 
To evaluate the generalization capability of the model, we test on the MPI-INF-3DHP dataset using the MPJPE, PCK, and AUC metrics. Real 2D pose estimates are used as input. As shown in Table \ref{tab:table_3dhp}, the Ours-large method achieves the best results in PCK, AUC, and MPJPE metrics, surpassing SOTA models with lower computational cost. The Ours-tiny method also achieves competitive results in MPJPE with an extremely small computational budget. These findings demonstrate that our approach is robust and performs well in both indoor and outdoor scenarios.

\subsection{Qualitative Analysis}

\begin{figure*}[ht]
	\centering
	\includegraphics[width=\linewidth]{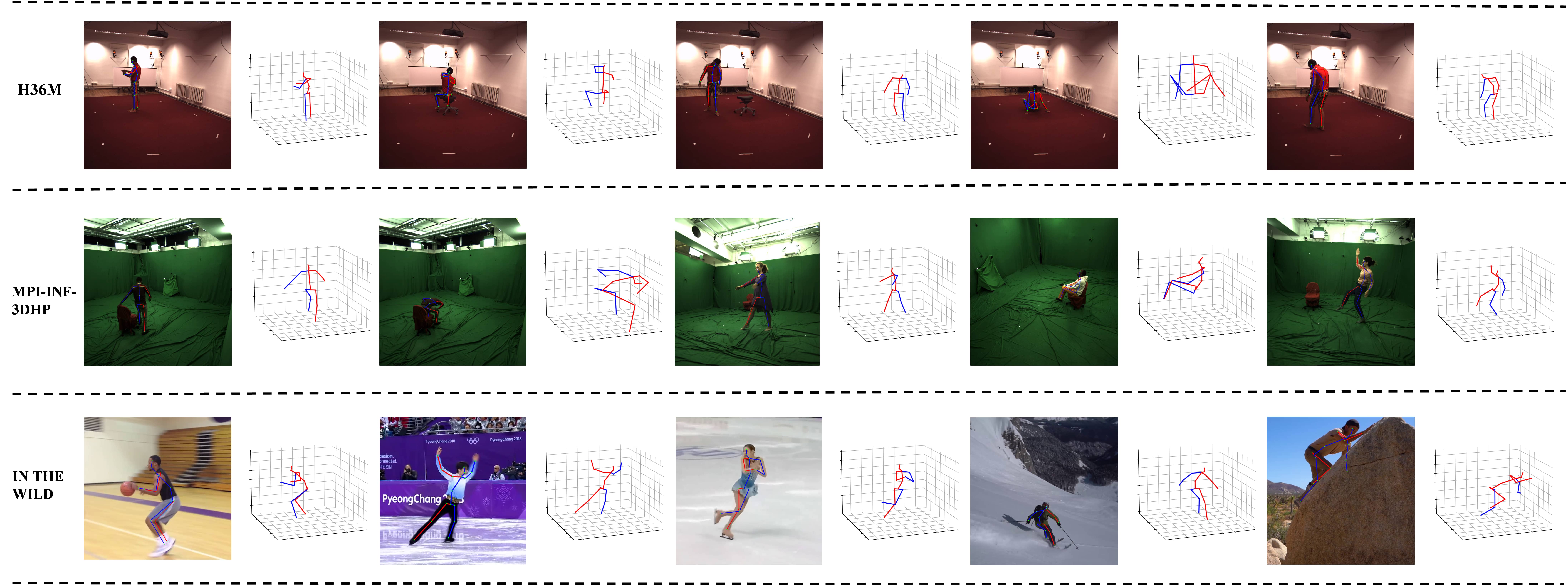}
	\caption{Visualization Results of Our Method on the Human3.6M Dataset, MPI-INF-3DHP Dataset, and In-the-Wild Videos}
	\label{fig:vis}
\end{figure*}

\begin{figure*}[ht]
	\centering
	\includegraphics[width=\linewidth]{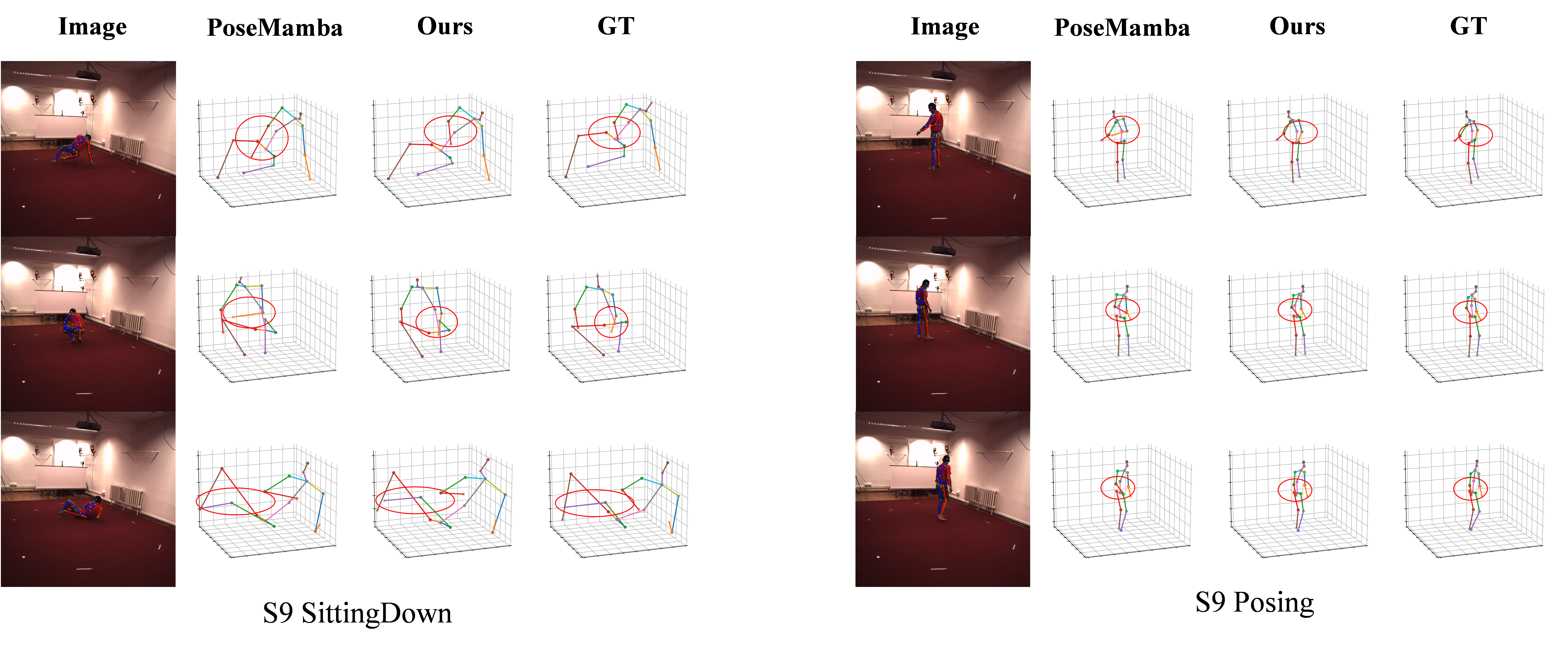}
	\caption{Visualization Comparison of Ours-large and PoseMamba Using 2D Poses with Erroneous Estimations.}
	\label{fig:vismixstecompare}
\end{figure*}
We further present the visualization results of our method in Figure \ref{fig:vis}. We showcase various actions from the Human3.6M dataset on test subject S9, such as phoning, sitting, and sitting down. Additionally, actions performed by subjects from the MPI-INF-3DHP dataset are also displayed. Finally, we include pose estimation results on motion videos in the wild scenarios, such as skating, skiing, and rock climbing. It can be observed that our model provides accurate pose estimation results.

Figure \ref{fig:vismixstecompare} illustrates a comparison of joint positions between Ours-large and PoseMamba-B ~\cite{huang2024posemamba} on actions with significant occlusions. When the input 2D poses contain substantial errors, our method demonstrates greater robustness compared to PoseMamba-B.

\subsection{Ablation Studies}
In this section, we validate the effectiveness of the proposed modules on the Human3.6M dataset. The experiments are conducted using the Ours-tiny model.
\subsubsection{Analysis on All Modules}
\begin{table*}[ht]\large
	\caption{Ablation Study on Contribution Analysis of Proposed Modules for Human3.6M Dataset}
	\label{tab:ablationallmodules}
	\centering
	\resizebox{1\linewidth}{!}{
	\begin{tabular}{@{}l | cc@{}}
		\toprule\toprule
		Module & MPJPE(P1) & MPJPE(P2) \\
		\midrule
		Baseline 																					&59.2   	 & 46.9 \\
		VIM  																 						&44.0        & 36.6         \\
		VIM + GEM(Spatiotemporal Refinement Module)  								 				&42.9        & 36.0         \\
		Spatiotemporal Refinement Module + Bone Aware Module 										&42.5        & 35.4         \\
		Spatiotemporal Refinement Module + Bone Aware Module + Bone-Joint Fusion Embedding			&41.7        & 34.8         \\
		\bottomrule\bottomrule
	\end{tabular}
	}
	\vspace{0.1cm}
\end{table*}

\begin{figure}[ht]
	\centering
	\includegraphics[width=\linewidth]{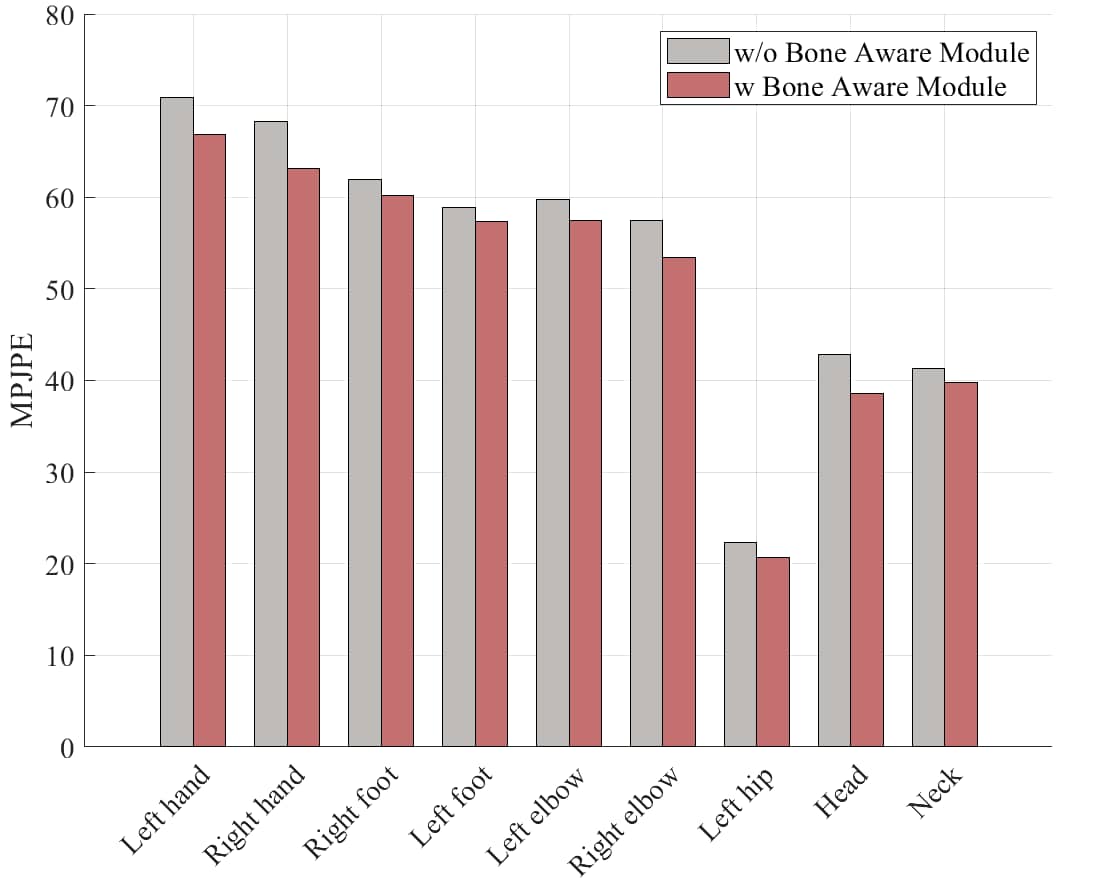}
	\caption{Comparison of Joint Errors Before and After Adding the Bone Aware Module}
	\label{fig:jointcompare}
\end{figure}

\begin{figure}[ht]
	\centering
	\includegraphics[width=\linewidth]{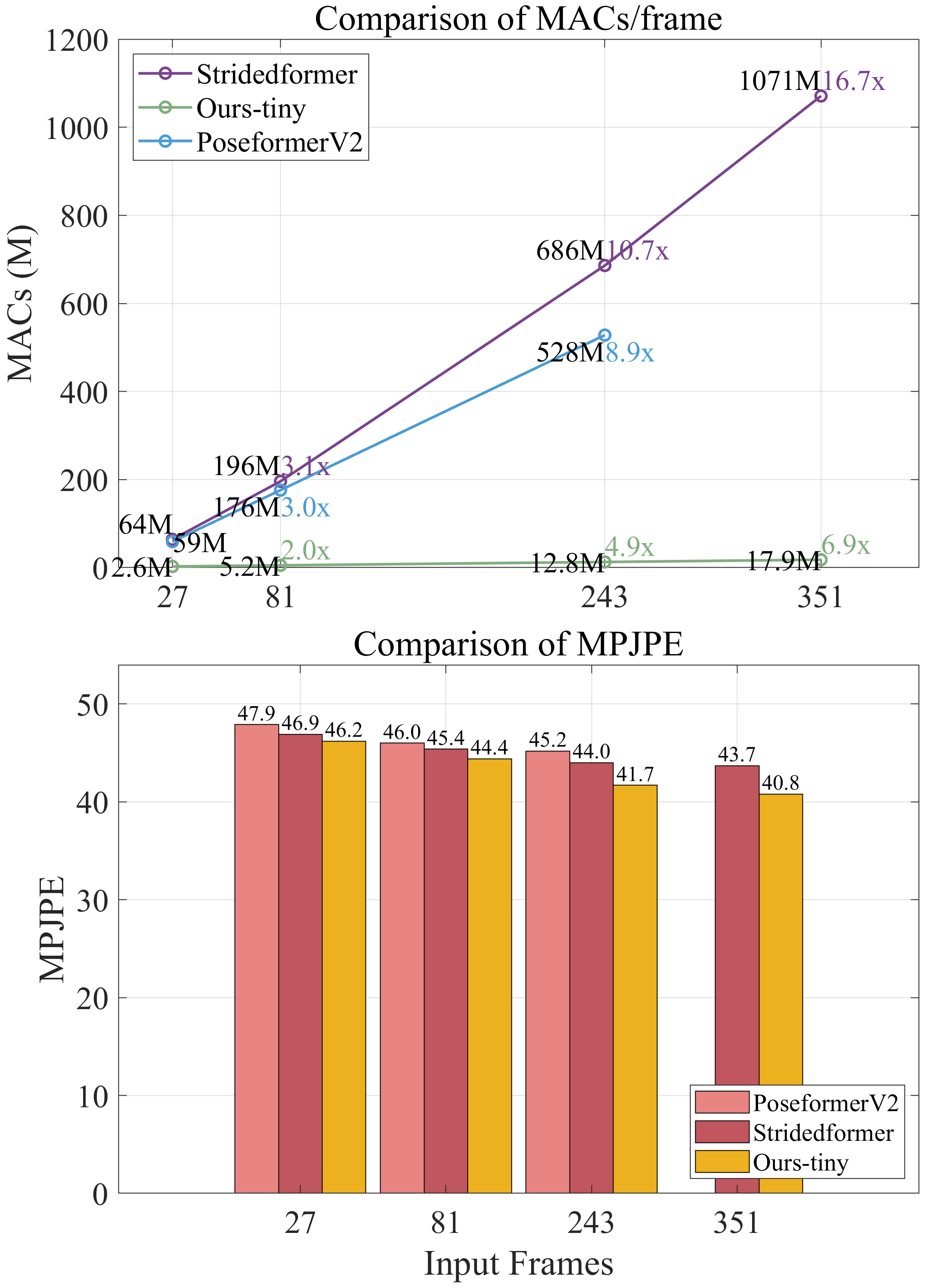}
	\caption{The comparison of computational cost (upper) and MPJPE (lower) between Ours-Tiny and StridedFormer with respect to different input sequence lengths. In the MACs comparison, \(m\times\) indicates the multiplier of the computational cost for the current sequence length relative to that of a 27-frame sequence. For example, \(10.7\times\) implies that the computational cost of the 243-frame model (\(686M\)) is 10.7 times that of the 27-frame model (\(64M\)).}
	\label{fig:mambastridedformer}
\end{figure}

We evaluated the effectiveness of various modules, including the VIM, GEM, bone aware Module, and bone-joint fusion embedding, as presented in Table \ref{tab:ablationallmodules}. The baseline model replaces the basic blocks in the spatiotemporal refinement module with Transformer encoder blocks, setting the depth to 8 and the hidden state dimension to 64, while keeping the remaining parameter settings consistent with the spatiotemporal refinement module. It can be observed that when the Transformer blocks are replaced with VIM blocks of the same configuration, the error decreases significantly. This is because previous Transformer-based pose estimators heavily relied on high-dimensional features, whereas we have set the feature dimension to 64. In contrast, VIM blocks are more suitable for low-dimensional features, significantly reducing computational cost while maintaining superior accuracy compared to Transformer blocks (44.0 mm vs. 59.2 mm). When the VIM blocks for processing spatial features are replaced with GEM blocks, the model (i.e., our spatiotemporal refinement module) achieves a further 1.0 mm reduction in MPJPE. This demonstrates that, building on the modeling capability of the Mamba model, our convolutional structure enhanced with graph convolution more effectively captures the spatial dependencies among joints. By directly concatenating the outputs of the bone aware module with the 2D pose and projecting them into a high-dimensional space through a linear layer, the MPJPE only decreases by 0.5 mm. However, employing the proposed bone-joint fusion embedding to learn representations from both coordinate systems before fusion proves more effective in guiding subsequent models to capture joint positions, resulting in a 1.3mm reduction in MPJPE.

The comparison of joint errors before and after adding the bone aware module is shown in Figure \ref{fig:jointcompare}. It can be observed that after incorporating the bone aware module, the joint errors for high-degree-of-freedom joints, such as the head, hands, feet, and knees, which are farther from the hip joint, have decreased. To evaluate the sensitivity of the proposed model to pose sequence length, we compared its performance with Stridedformer \cite{li2022exploiting} and PoseformerV2 \cite{zhao2023poseformerv2} using input sequences of varying lengths. As shown in Figure \ref{fig:mambastridedformer}, Ours-tiny achieves a significant advantage in computational efficiency. As the input sequence length increases, although both Stridedformer and PoseFormerV2 incorporate modules for accelerated computation, their computational costs still surge dramatically. In contrast, the computational cost curve of Ours-tiny remains relatively flat. Furthermore, even with a significantly lower computational cost, our method consistently delivers superior performance.

\subsubsection{Bone Aware Module}
\begin{table}[ht]\Large
	\caption{Ablation study on the bone aware module. Categories represent the number of classification categories, and ACC denotes the accuracy of the classification results. Coordinate Systems indicate the representation method of the generated bone vectors within the coordinate system.}
	\label{tab:boneaware}
	\centering
	\resizebox{1\linewidth}{!}{
		\begin{tabular}{@{}l | cc@{}}
			\toprule\toprule
			Categories 																&ACC  		 &MPJPE(P1) 	\\
			\midrule
			2																		&91\%		 & 43.9			\\
			4																 		&85\%        & 42.8         \\
			6 																		&77\%        & 41.7         \\
			7							 											&76\%        & 42.5         \\
			8																		&71\%		 & 42.4         \\
			\bottomrule
			\toprule
			Coordinate Systems														&MPJPE(P1)	&MPJPE(P2)		\\
			\midrule
			Bone vector in Cartesian coordinate										&42.1		&35.0			\\
			Bone vector in Spherical coordinates									&41.7		&34.8			\\
			\bottomrule\bottomrule
		\end{tabular}
	}
	\vspace{0.1cm}
\end{table}
\begin{table}[ht]\normalsize
	\caption{The network architecture of the Bone Aware Module is illustrated, where \(Depth\) denotes the network depth, \(Dim\) the hidden feature dimension.}
	\label{tab:bamdepthdim}
	\centering
	\begin{tabular}{@{}lc | ccc@{}}
		\toprule\toprule
		Depth	&Dimension   &Param  &MACs  &Accuracy 	\\
		\midrule
		3		&64		 	 &107.57K 	&199M &0.76 \\
		4		&64          &110.53K 	&202M &0.77  \\
		5 		&64          &111.75K   &203M &0.77  \\
		3		&128         &376.17K	&775M &0.78   \\
		4		&128		 &384.12K  	&786M &0.78   \\
		5		&128		 &387.07K	&790M &0.78    \\
		\bottomrule\bottomrule
	\end{tabular}
	\vspace{0.1cm}
\end{table}
%\begin{table}[ht]\Large
%	\caption{Ablation study on the fixed value types for angle categories.}
%	\label{tab:representation}
%	\centering
%	\resizebox{1\linewidth}{!}{
	%		\begin{tabular}{@{}l | cc@{}}
		%			\toprule\toprule
		%			Angle Representation													&MPJPE(P1)	&MPJPE(P2)		\\
		%			\midrule
		%			Normalization															&43.6		&35.8		\\
		%			Midpoint of the interval												&41.7		&34.8		\\
		%			One-third point of the interval											&42.2		&34.9		\\
		%			Two-thirds point of the interval										&42.3		&35.4		\\
		%			One-quarter point of the interval										&43.3		&35.7		\\
		%			\bottomrule\bottomrule
		%		\end{tabular}
	%	}
%	\vspace{0.1cm}
%\end{table}
The ablation study on the bone aware module is using detected 2D poses as input. First, we analyze the impact of the number of predicted polar angle categories in the bone aware module. As shown in the upper section of Table \ref{tab:boneaware}, an ablation study is performed across various category settings. When the classification is limited to two categories, the bone aware module achieves the highest classification accuracy of 91\%; however, this configuration also results in the largest error. This is because classifying into only two categories provides only two directional cues for bone depth, which is overly coarse and fails to offer meaningful guidance for the model. As the number of categories increases, the bone aware module provides finer directional guidance, but the classification accuracy decreases accordingly. The best MPJPE is achieved with six categories. Additionally, we evaluate the coordinate representation of the output bone vectors. As shown in the lower section of Table \ref{tab:boneaware}, spherical coordinates perform better than Cartesian coordinates because they incorporate explicit information about bone angles and lengths.

Table \ref{tab:bamdepthdim} presents the impact of various network configurations on prediction accuracy. The results indicate that the performance gain from increasing the model size is marginal. Therefore, selecting a configuration with a depth of 4 and a hidden dimension of 64 yields reasonably accurate classification results with minimal computational cost.

%The ablation study on the fixed value representations for polar angle classification results is shown in Table \ref{tab:representation}. Normalization represents mapping the classification results to normalized values within the range of \(0\) to \(\pi\), representing the polar angle in radians. Midpoint refers to using the midpoint of the radian range corresponding to each category, such as \(\frac{1}{12}\pi\) for the first category. Similarly, one-third point, two-thirds point, and one-quarter point represent the respective points within the radian range of each category. These values provide diverse representations; however, we observe that using the midpoint achieves the smallest error. This is because the midpoint better represents the average radian for the entire category, leading to more accurate guidance for the model.

\subsubsection{Spatiotemporal Refinement Module}
\begin{table}[ht]
	\caption{Ablation study on the fusion positions of GCN and Mamba modules. \(Parallel\) denotes parallel processing streams of GCN and Mamba. \(Sequential(GCN+Mamba)\) denotes the configuration where the input is first processed by GCN followed by Mamba, whereas \(Sequential(Mamba+GCN)\) denotes the reverse order. \(Inner\) indicates that GCN is embedded within the internal structure of the Mamba block.}
	\label{tab:gemstructure}
	\centering
	\resizebox{1\linewidth}{!}{
		\begin{tabular}{@{}l | ccc@{}}
			\toprule\toprule
			Structure 																		&Param  &MACs	&MPJPE(P1)  \\
			\midrule
			Parallel																		&765.6K &2.6G &43.8	\\
			Sequential(GCN+Mamba)															&848.8K &2.8G &44.1 \\
			Sequential(Mamba+GCN)															&848.8K &2.8G &44.0 \\
			Inner																			&748.1K &2.6G &42.9\\
			\bottomrule\bottomrule
		\end{tabular}
	}
	\vspace{0.1cm}
\end{table}

\begin{table}[ht]\large
	\caption{Ablation study on scanning direction and basic blocks in the spatiotemporal refinement module. To verify the effectiveness of bidirectional GEM under similar computational costs, \(\dag\) indicates an increase in model scale.}
	\label{tab:ablationSpatiotemporal}
	\centering
	\resizebox{1\linewidth}{!}{
		\begin{tabular}{@{}l | ccc@{}}
			\toprule\toprule
			Module 																		&Param  &MACs	&MPJPE(P1)  \\
			\midrule
			Unidirectional w. VIM														&564.3K	&1.6G	&44.2       \\
			Bidirectional w. VIM 														&612.4K	&1.9G	&44.0       \\
			Unidirectional w. (VIM + GEM)							 					&700.0K	&1.9G	&43.4       \\
			Bidirectional w. VIM\dag 													&870.6K	&2.5G	&43.2		\\
			Bidirectional w. (VIM + GEM)												&748.1K	&2.6G	&42.9		\\
			\bottomrule\bottomrule
		\end{tabular}
	}
	\vspace{0.1cm}
\end{table}

\begin{figure}[ht]
	\centering
	\includegraphics[width=\linewidth]{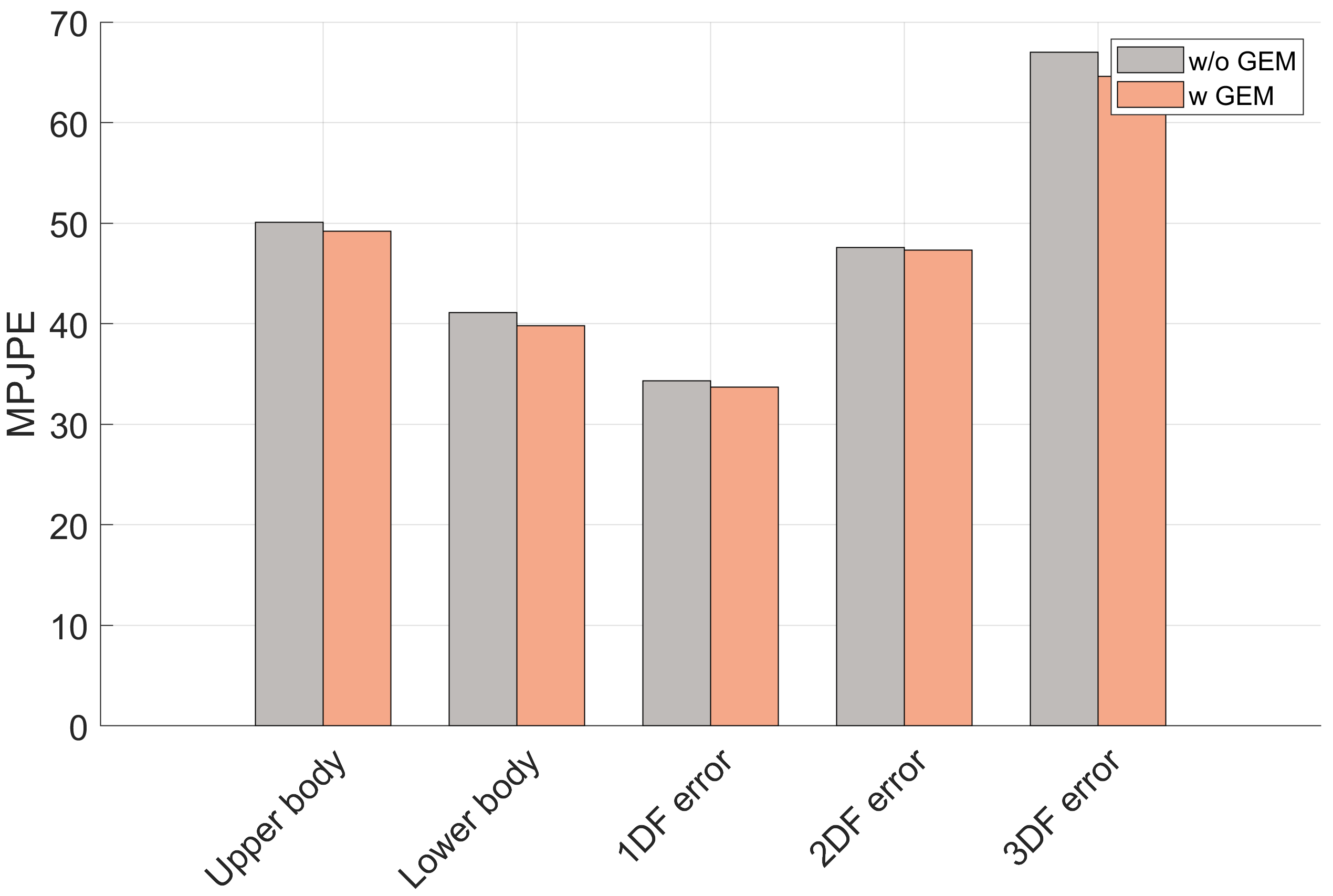}
	\caption{Comparison of Joint Errors for Different Body Parts and Degrees of Freedom Before and After GEM Module Integration}
	\label{fig:gemerror}
\end{figure}
In Table \ref{tab:gemstructure}, we evaluate the effectiveness of different integration strategies between GCN and the Mamba model. The Parallel design processes the GCN and Mamba streams independently and fuses the outputs of the two branches. The Sequential design represents the sequential connection of GCN and Mamba. Inner, GEM's structure, embeds GCN before the 1D convolutional structure within the Mamba model. The results indicate that directly feeding the structural information output by GCN into 1D convolution is more beneficial for Mamba to capture local relationships. This architectural effectively alleviates the Mamba's original insensitivity to human body topological structures.

We also conducted ablation experiments on the directions of VIM and GEM within the spatiotemporal refinement module. The results are reported in Table \ref{tab:ablationSpatiotemporal}. When only the VIM block is used with an undirected scanning pose sequence, the error reaches a maximum of 44.2 mm. By adopting a bidirectional scanning approach, the error decreases to 44.0 mm. Replacing the VIM module with GEM to handle the spatial joint relationships reduces the MPJPE to 43.4 mm for the unidirectional model. Adding a bidirectional graph convolution structure achieves the lowest MPJPE of 42.9 mm. Additionally, we increased the depth of the bidirectional VIM (denoted as \(\dag\)) to achieve a similar computational cost to the bidirectional graph convolution. The better results (43.2mm vs. 42.9mm) demonstrate that the effectiveness of the GEM module cannot be attributed solely to the increase in computational cost. The inclusion of the GEM module demonstrates its effectiveness in helping the Mamba model learn the local relationships of joints. 

The comparison of errors before and after adding the GEM module to the bidirectional Mamba network is shown in Figure \ref{fig:gemerror}. We calculate the joint errors for the upper body, lower body, and joints with different degrees of freedom. It can be observed that the errors for various body parts have decreased, with the largest reduction in the third-degree-of-freedom joints. This demonstrates the effectiveness of the GEM module in predicting high-degree-of-freedom joints.

\section{Conclusion}
In this paper, we propose a Mamba-Driven Topology Fusion framework for monocular 3D human pose estimation. We first design a bone aware module to estimate the direction and position of bone vectors, providing topological guidance for the Mamba model in capturing dependencies within joint sequences. Based on the Vision Mamba block, we further introduce GEM, which incorporates internal forward and backward Graph Convolutional Networks to enhance Mamba’s ability to capture local joint relationships. Finally, we propose a bone-joint fusion embedding to integrate joint and bone features across different coordinate systems, along with a spatiotemporal refinement module for accurate 3D human pose regression. Our method achieves superior results on the Human3.6M and MPI-INF-3DHP datasets compared to previous methods, achieving a win-win in both speed and accuracy.

\bibliographystyle{IEEEtran}
\bibliography{references.bib}

\end{document}